\pdfoutput=1

\documentclass[11pt]{article}

\usepackage{ACL2023}

\usepackage{times}
\usepackage{latexsym}

\usepackage[T1]{fontenc}

\usepackage[utf8]{inputenc}

\usepackage{microtype}

\usepackage{inconsolata}

\title{Optimizing Abstractive Summarization With Fine-Tuned
PEGASUS}

\author{
Sadiul Arefin Rafi \and Naimur Rahman \and Kazi Nazibul Islam \and Ha-mim Ahmad \\
BRAC University \\
\texttt{\{sadiul.arefin.rafi, naimur.rahman1, kazi.nazibul.islam, ha.mim.ahmad\}@g.bracu.ac.bd}
\AND
Farig Yousuf Sadeque \\
BRAC University \\
Associate Professor \\
\texttt{farig.sadeque@bracu.ac.bd}
}

\begin{document}
\maketitle
\begin{abstract}
Abstractive text summarization is the technique of generating a short and concise summary comprising the salient ideas of a source text without making a subset of the salient sentences from the source text. The introduction of transformer models such as BART, T5, and PEGASUS has made this sort of summarization process more efficient and accurate. The objective of this paper is to fine-tune PEGASUS on the XL-Sum English corpus to achieve a better performance compared to the baseline mT5 model. The performance of the generated summaries from the fine-tuned model is evaluated using the ROUGE metric, which basically compares the auto-generated summaries with human-created summaries. To the best of our knowledge, the results from our fine-tuned PEGASUS model give a state-of-the-art performance on the XL-Sum English Corpus. To quantify the improvement, there is a 4.04\% improvement in the ROUGE-1 score, a 15.25\% increase in the ROUGE-2 score, and a 3.39\% improvement in the ROUGE-L score from the baseline model.
\end{abstract}

\section{Introduction}

In our data-rich world, written content saturates every aspect of contemporary culture, spanning news journalism, scholarly research, social media, and product reviews. The deluge of daily textual data necessitates automated text summarization methods, a core component of Natural Language Processing (NLP) \cite{sankarasubramaniam}. These methods distill and present textual content concisely, empowering readers to swiftly grasp crucial information, transcending the need to plow through voluminous source documents. Text summarization stands as a cornerstone in information retrieval and comprehension, poised to reshape our navigation of the expanding sea of knowledge.

Extractive summarization and abstractive summarization are two broad categories of text summarizing approaches. In order to create the summary, extractive summarization chooses a subset of important sentences from the original text. In contrast, abstractive summarization produces a unique summary that captures the key points without necessarily recycling entire passages from the source material \cite{Abdel-Salam}. Although employing pre-existing sentences simplifies extraction approaches, they frequently fail to preserve coherence and capture the entire context. While producing summaries that are more coherent and contextually accurate, abstractive approaches nevertheless add more complexity by necessitating the maintenance of factual integrity and the resolution of ambiguities.

The increased need for advanced NLP systems has sparked research projects to determine how well pre-trained language models like PEGASUS can be tuned to enhance abstractive summarization. Traditional sequence-to-sequence architecture-based attention-enhanced abstractive summarization models have demonstrated promising results. They do have certain limitations, though, including the occasional use of redundant terminology and concerns with variable quality and terms that are not commonly used. Sequence-to-sequence models' slowness in concurrent data processing is one of their significant drawbacks. The self-attention mechanism built into the transformer architecture, which enables parallel computing over the whole input sequence, mitigates this \cite{zhang2020pegasus}. 

In our research, we fine-tuned the PEGASUS transformer model utilizing the XL-Sum corpus to optimize its performance.

\section{Research Objective}

The core objectives of this research are to explore, develop, and evaluate transformer models for abstractive text summarization. The specific goals are as follows:

\begin{itemize}
    \item To identify an efficient transformer model that is apt for abstractive text summarization, providing a balance between computational efficiency and summarization quality.
    
    \item To fine-tune the identified efficient transformer model on the XL-Sum corpus to adapt its parameters and specialize its functionality to the characteristics of the dataset.
    
    \item To develop a state-of-the-art model that demonstrates superior performance in abstractive summarization tasks compared to the baseline model, mT5.
    
    \item To rigorously compare the ROUGE scores of the fine-tuned model with the baseline model mT5, to quantify the enhancement in performance and summarization quality.
\end{itemize}

\section{Literature Review}

In 2020, Pilault et al. presented a method employing neural abstractive summarization to generate abstractive summaries of extensive texts, often exceeding a few thousand words. Their approach involves an initial extractive step to train the transformer language model on relevant data before it produces a summary. Notably, this technique achieves superior ROUGE scores and yields more abstractive summaries in comparison to previous methodologies that utilized a copy mechanism \cite{pilault2020extractive}.\\

Ensuring the accuracy of information between the produced summary and the original content poses a significant challenge in the field of abstractive summarization. Contemporary models that have been trained on preexisting datasets exhibit a phenomenon known as entity hallucination, wherein they generate references to entities that do not actually exist within the original text. One potential solution to address the issue of entity hallucination is to apply a filtering process to the training data and incorporate supplementary metrics for assessing the factual consistency of the summaries \cite{nan2021entity}.\\

The PEGASUS model is a sequence-to-sequence model optimized for abstractive text summarization through the use of gap-sentence generation as a pre-training objective. Similar to an extractive summary, PEGASUS takes an input document and masks important sentences before generating a single output sequence from the remaining text. In terms of selecting a pre-training corpus, C4 and HugeNews were considered for training the $PEGASUS_{LARGE}$ model on them. For downstream summarization, 12 public abstractive summarization datasets were used (Zhang et al., 2019). For example, XSum, CNN, NEWSROOM, Gigaword, and so on. Initially, a reduced-size model with 223 million parameters, $PEGASUS_{BASE}$ was conducted by using 4 out of 12 datasets for faster computation. Then the pre-training was scaled up by introducing $PEGASUS_{LARGE}$ with 568 million parameters, which used all 12 datasets. Finally, the result showed that $PEGASUS_{BASE}$ and $PEGASUS_{LARGE}$ had huge performance improvements on downstream datasets. $PEGASUS_{BASE}$ was able to achieve SOT on a number of datasets, but $PEGASUS_{LARGE}$ was successful in exceeding SOT on all downstream datasets \cite{zhang2020pegasus}.

\section{Description of XL-Sum Corpus}
An enormous dataset called XL-Sum was created especially for the job of abstractive summarization. It stands out for its diversity; it includes more than a million pairs of articles with expert annotations and their summaries, all taken from the BBC. Language diversity is one of its unique characteristics; XL-Sum offers information in 44 different languages, some of which lack widely accessible datasets. It is therefore extremely useful for research and other applications that need multilingual capabilities.\\

The dataset performs exceptionally well in a number of domains, according to both human assessments and intrinsic measures. It enables summaries that are both succinct and abstract while maintaining a high standard of quality. In general, XL-Sum is a reliable tool for creating and refining abstractive summarization models \cite{hasan-etal-2021-xl}.\\

\subsection{Variations}
There are two versions of the XL-Sum dataset. The newer version of the dataset includes the Traditional Chinese language. By adding this language, the XL-sum dataset enables better formatting, better extraction, larger evaluation splits, and more data. With the inclusion of the Traditional Chinese language, XL-Sum became the largest text summarization dataset publicly available \cite{hasan-etal-2021-xl}.

\subsection{Split}

\begin{center}
\begin{tabular}{ |c|c|c|c| } 
 \hline
 \textbf{Language} & \textbf{Train} & \textbf{Dev} & \textbf{Test}\\ 
 \hline
 Most other languages & 80\% & 10\% & 10\%\\ 
 \hline
 English & 93\% & 3.5\% & 3.5\%\\ 
 \hline
\end{tabular}
\end{center}
\vspace{1em}

\section{Fine-tuning PEGASUS on XL-Sum}
The PEGASUS transformer model, which Google AI Research initially developed for abstractive summarization, is the subject of our empirical study in this chapter. Our main focus is on the improvements made possible by perfecting this model on the XL-Sum dataset. Through our efforts, we were able to create a model that, when tested on the same dataset, performs better than the mT5 baseline. Particularly, our improved PEGASUS model demonstrated significantly enhanced performance in ROUGE-1, ROUGE-2, and ROUGE-L measures.

\subsection{Fine-Tuning the PEGASUS Transformer Model}
A large percentage of our study and investigation is focused on fine-tuning. For this procedure, we used 20\% of the English training corpus from XL-Sum and the following hyperparameters:

\begin{itemize}
    \item Learning rate: 2e-05
    \item Training batch-size: 8
    \item Evaluation batch-size: 8
    \item Seed: 42
    \item Optimizer: Adam with betas=(0.9,0.999) and epsilon=1e-08
    \item LR scheduler type: linear
    \item Number of epochs: 5
\end{itemize}

It is crucial to remember that these hyperparameters were chosen following a number of tests and parameter sweeps to enhance performance. These decisions aided in our model's convergence and led to the production of better results.

\section{Comparative Evaluation: PEGASUS vs. mT5}

As a baseline model for comparison, we used the mT5 model fine-tuned on XL-Sum corpus which was released by Hasan et al. Its ROUGE scores for ROUGE-1, ROUGE-2, and ROUGE-L were 37.60, 15.15, and 29.88, respectively \cite{hasan-etal-2021-xl}.

\begin{table}[!ht]
\small
\centering
\setlength{\tabcolsep}{6pt} 
\begin{tabular}{ |l|c|c|c| }
\hline
Model & ROUGE-1 & ROUGE-2 & ROUGE-L \\
\hline
 Baseline mT5 & 37.601 & 15.153 & 29.88 \\
\hline
PEGASUS XLSum & \textbf{39.121} & \textbf{17.467} & \textbf{30.894} \\
\hline
\end{tabular}
\caption{Rouge scores from the mT5 and fine-tuned PEGASUS}
\label{table:rouge_scores}
\end{table}

These results show a considerable performance gain when compared to our improved PEGASUS model. Our model achieved ROUGE scores of 39.121 for ROUGE-1, 17.467 for ROUGE-2, and 30.894 for ROUGE-L, respectively. The improvement in these measures illustrates the efficacy of our process of fine-tuning and confirms the potency of the PEGASUS transformer model when appropriately suited for certain workloads.

\section{Conclusion}

In this work, we experimented with various transformer architectures to fine-tune abstractive text summarization models on the XL-Sum English dataset. The mT5\_multilingual\_XLSum model by Hasan et al. presents ROUGE scores of 37.601 for ROUGE-1, 15.153 for ROUGE-2, and 29.88 for ROUGE-L. However, significant improvements were observed when we used the pre-trained PEGASUS model for fine-tuning. We named our developed model pegasus\_xlsum, which outperformed the mT5\_multilingual\_XLSum model in all ROUGE metrics, achieving ROUGE-1, ROUGE-2, and ROUGE-L scores of 39.121, 17.467, and 30.894, respectively. To quantify the improvement, we could see a \textbf{4.04\%} improvement in the ROUGE-1 score and a \textbf{3.39\%} improvement in the ROUGE-L score. Our model exhibited a notable improvement, showing a \textbf{15.25\%} increase in the ROUGE-2 score, which indicates a substantial enhancement in capturing more complex sentence structures and content relationships. These results show our fine-tuned model not only outperformed the baseline model mT5, fine-tuned on the XL-Sum corpus but also achieved state-of-the-art performance on its English corpus.

\section{Limitations}

Transformer models, which are large pre-trained models, are computationally complex and require significant hardware resources to run efficiently. While fine-tuning the baseline model, we implemented early stopping to terminate training after a predetermined number of epochs if the model's performance on the evaluation dataset does not improve. As convention dictates, a higher number of epochs should be set for fine-tuning. However, due to the constraints, we faced in terms of resources, we decided to fine-tune it on 20\% of the training corpus and terminate the training after only five epochs. Another limitation of our fine-tuned transformer model is the ability to deal with input text with short lengths. It was discovered that in order to produce a coherent and concise summary, a minimum of several lines of input text was necessary.

\bibliographystyle{acl_natbib}
\bibliography{custom}
\appendix

\end{document}